# AxonDeepSeg: automatic axon and myelin segmentation from microscopy data using convolutional neural networks

**Aldo Zaimi[+,1], Maxime Wabartha[+,1,2], Victor Herman[1,2], Pierre-Louis Antonsanti[1,3], Christian S. Perone[1], Julien Cohen-Adad[1,4].**

(+) These authors contributed equally to this work.

(1) NeuroPoly Lab, Institute of Biomedical Engineering, Polytechnique Montreal, Montreal, QC, Canada
(2) Ecole Centrale de Lille, Lille, France
(3) Ecole Centrale de Nantes, Nantes, France
(4) Functional Neuroimaging Unit, CRIUGM, Université de Montréal, Montreal, QC, Canada

## Abstract

Segmentation of axon and myelin from microscopy images of the nervous system provides useful quantitative information about the tissue microstructure, such as axon density and myelin thickness. This could be used for instance to document cell morphometry across species, or to validate novel non-invasive quantitative magnetic resonance imaging techniques. Most currently-available segmentation algorithms are based on standard image processing and usually require multiple processing steps and/or parameter tuning by the user to adapt to different modalities. Moreover, only few methods are publicly available. We introduce *AxonDeepSeg*, an open-source software that performs axon and myelin segmentation of microscopic images using deep learning. *AxonDeepSeg* features: (i) a convolutional neural network architecture; (ii) an easy training procedure to generate new models based on manually-labelled data and (iii) two ready-to-use models trained from scanning electron microscopy (SEM) and transmission electron microscopy (TEM). Results show high pixel-wise accuracy across various species: 85% on rat SEM, 81% on human SEM, 95% on mice TEM and 84% on macaque TEM. Segmentation of a full rat spinal cord slice is computed and morphological metrics are extracted and compared against the literature. *AxonDeepSeg* is freely available at https://github.com/neuropoly/axondeepseg.

## Introduction

Neuronal communication is ensured by the transmission of action potentials along white matter axons. For long distance communication, these axons, which are typically 1-10μm in diameter, are surrounded by a myelin sheath whose main role is to facilitate the propagation of the electrical impulses along neuronal fibers and increase the transmission speed[1,2]. Pathologies such as neurodegenerative diseases (e.g., multiple sclerosis) or trauma are associated with myelin degeneration, which can ultimately lead to sensory and motor deficits (e.g., paraplegia)[3,4]. Being able to image axons and myelin sheaths at high resolution would help researchers understand the origins of demyelination and test therapeutic drugs[5,6] and could also be used to validate novel magnetic resonance imaging biomarkers of myelin[7]. High resolution histology is typically done using electron microscopy following osmium staining to obtain myelin contrast. Then, axons and myelin can be analysed on the images to derive metrics such as axon density or myelin thickness. However, given that 1 mm² of white matter can contain over 100,000 axons, it is important to obtain a robust and reliable segmentation of individual axons and myelin as automatically as possible.

Several segmentation methods for axon and myelin have been proposed which are based on traditional image processing algorithms including thresholding and morphological operations[8,9], axon shape-based morphological discrimination[10], watershed[11,12], region growing[13], active contours without[14,15] and with discriminant analysis[15]. However, few limitations can be reported from the previous work: (i) traditional image-based methods are designed to work on specific imaging modalities and often fail if another contrast is used (e.g., optical image instead of electron microscopy); (ii) previous methods are not fully-automatic as they typically require either preprocessing, hand-selected features for axon discrimination and/or postprocessing; (iii) traditional image-based methods do not make full use of the contextual information of the image (i.e., multi-scale representation of axons, average shape of axons, etc.) and (iv) most of the previous methods are not publicly available (to our knowledge, only that from[14,15] are).

In the last five years, deep learning methods have become the state of the art when it comes to computer vision tasks. Convolutional neural networks (CNNs) are particularly suited to image classification[16–19] and semantic segmentation[20]. Cell segmentation is one of the popular application of CNNs[21,22]. The U-Net architecture introduced by Ronneberger and collaborators[23] has inspired many medical segmentation applications, efficiently combining both context and localization of structures of interest. Segmentation of axons and myelin based on deep learning approaches offers significant advantages when compared with traditional image segmentation algorithms: (i) there is no need to hand-select relevant features because the network is able to learn the hidden structural and textural features by itself, (ii) this approach allows to segment both axons and myelin sheaths in two different labels with the same network, without the need of any explicit pre- or post-processing, (iii) the network can be trained for various imaging modalities without significantly changing its architecture and (iv) once trained, the model is relatively fast at the prediction step (a few seconds) compared to more traditional image processing methods.

Few research groups have applied deep learning for axon and myelin segmentation. Naito and collaborators[24] have implemented a two-step process that first performs clustering segmentation of myelinated nerve fibers in optical microscopic images, and then discriminates between true and false candidates by using a CNN classification network. This group did not exploit the CNN for the segmentation, but only for discrimination. The work from Mesbah and collaborators[25] presented a deep encoder-decoder CNN that can segment both axon and myelin and claimed to achieve up to 82% pixel-wise accuracy. However, the network has been designed specifically for light microscopy images, the implementation is not publicly available and minimal regularization strategies have been employed in order to improve generalization.

We present *AxonDeepSeg*, a deep learning framework for robust and automatic segmentation of both axons and myelin sheaths in myelinated fibers. *AxonDeepSeg* features: (i) a CNN architecture for semantic segmentation of histological images; (ii) two ready-to-use models for the segmentation of scanning electron microscopy (SEM) and transmission electron microscopy (TEM) samples adapted to a variety of species and acquisition parameters; (iii) a well-documented training pipeline to generate models for new imaging modalities and (iv) free and open source code (https://github.com/neuropoly/axondeepseg).

This paper is organized as follows. The Methods section lists the datasets used, details the architecture of the network and presents the validation methodology. The Results section presents axon and myelin segmentation results obtained on SEM and TEM samples, and shows an example application to extract morphological metrics from a full rat spinal cord slice. The Discussion section addresses the advantages and limitations of our models and discusses further possible improvements.

# Methods

## Dataset

Microscopy images used in this study were acquired with two different imaging techniques: SEM and TEM. Different acquisition resolutions were used, in order to increase variability and obtain better generalization of the model, with isotropic pixel size resolution ranging from 0.05 to 0.18 μm (SEM) and 0.002 to 0.009 μm (TEM). SEM samples were stained with 2% osmium, embedded in epoxy, polished and imaged with the same SEM system (Jeol 7600F). TEM images were obtained from mice brain samples (splenium), as described in[26]. Additionally, a macaque sample of the corpus callosum was added to the test set. Preparation and imaging procedures are described in[7]. Table 1 lists the samples used for the experiments.

All methods were carried out in accordance with relevant guidelines and regulations. Experimental protocols involving rats were approved by the Montreal Heart Institute committee. Experimental protocols involving the human spinal cord were done at the anatomy laboratory of the University of Quebec at Trois-Rivieres. The spinal cord donor gave informed consent and procedures were approved by the local ethics committee (SCELERA-15-03-pr01). Similarly, TEM images shared by collaborators were obtained in accordance with the corresponding ethics committees (mice: Institutional Animal Care and Use Committee at the New York University School of Medicine, macaque: Montreal Neurological Institute Animal Care Committee).

|  |  | Number of images | Species | Tissue | Pixel size (μm) | FOV (μm²) | Tissue preparation (% paraformaldehyde – % glutaraldehyde) |
|---|---|---|---|---|---|---|---|
| SEM | Training / validation | 1 | Rat | Spinal cord (cervical) | 0.18 | 230×166 | 4% – 2% |
|  |  | 3 | Rat | Spinal cord (cervical) | Between 0.05 and 0.17 | Between 132×90 and 218×162 | 4% – 0% |
|  |  | 3 | Rat | Spinal cord (cervical) | 0.1 | Between 74×76 and 77×84 | 3% – 3% |
|  |  | 1 | Rat | Spinal cord (cervical) | 0.13 | 247×234 | 3% – 3% |
|  |  | 1 | Rat | Spinal cord (cervical) | 0.1 | 82×77 | 3% – 3% |
|  | Testing | 1 | Rat | Spinal cord (cervical) | 0.13 | 150×97 | 3% – 3% |
|  |  | 1 | Rat | Spinal cord (cervical) | 0.07 | 108×77 | 3% – 3% |
|  |  | 1 | Human | Spinal cord (cervical) | 0.13 | 715×735 | 4% – 2% |
| TEM | Training / validation | 8 × 17 mice | Mouse | Brain (splenium) | 0.002 | 6×9 | 2% – 2.5% |
|  | Testing | 8 × 3 mice | Mouse | Brain (splenium) | 0.002 | 6×9 | 2% – 2.5% |
|  |  | 1 | Macaque | Brain (corpus callosum) | 0.009 | 27×21 | 2% – 2% |

**Table 1:** List of datasets used for the experiments. For each sample, the following information is indicated: number of images used, species, tissue type, pixel size, field of view (FOV) and tissue preparation details. For the scanning electron microscopy (SEM) model, training was done on rat spinal cord samples and testing was performed on rat and human spinal cord samples. For the transmission electron microscopy (TEM) model, training was done on mice brain samples and testing was performed on mice and macaque brain samples.

## Gold standard labelling

The gold standard labelling on SEM samples was created as follows: (i) Myelin sheaths were manually segmented (inner and outer contours) with GIMP (https://www.gimp.org/); (ii) Axon labels were obtained by filling the region enclosed by the inner border of the myelin sheaths; (iii) Small manual corrections were done on the axon and myelin masks (contour refinement, elimination of false positives) when necessary.

The gold standard labelling on TEM samples was created as follows: (i) Myelin was first segmented using intensity thresholding followed by manual correction, then the inner region was filled to generate axon labels. More details can be found about the generation of labels for the macaque[7] and the mice[26].

All gold standard labels were cross-checked by at least two researchers. The final gold standard consists of a single *png* image with values: background=0, myelin=127, axon=255. Example SEM and TEM samples and corresponding gold standard labels are shown in Figure 1. This figure also illustrates the large variability in terms of image features, especially for the SEM data (contrast, noise, sample preservation, etc.).

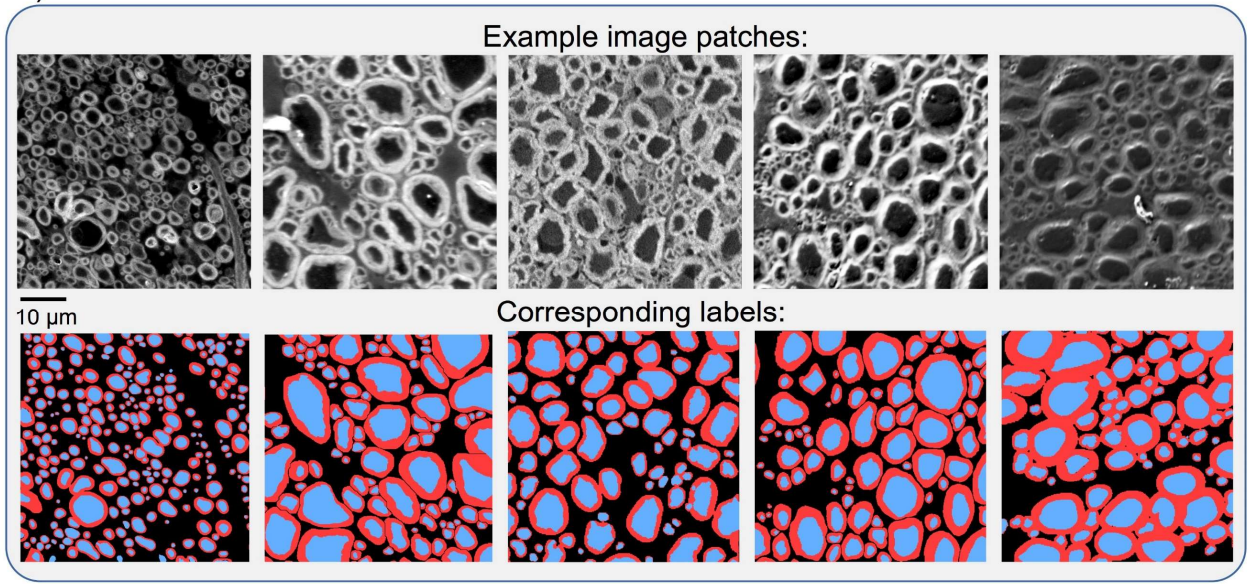

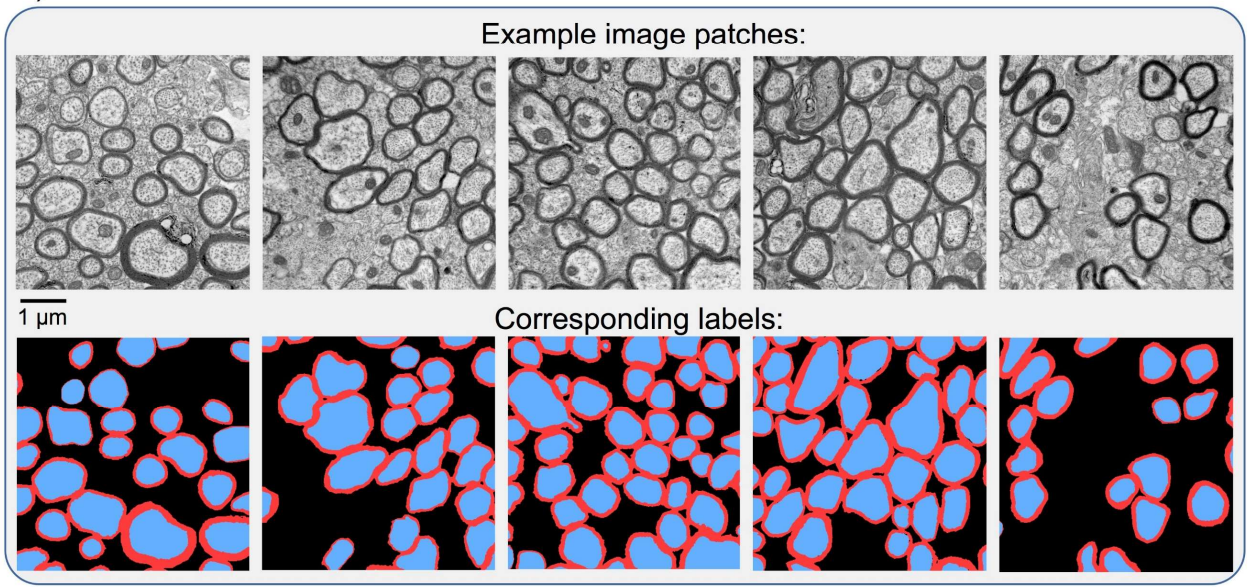

**Figure 1:** Overview of the data and gold standard labels for SEM (a) and TEM (b). Label masks contain 3 classes: axon (in blue in the figure), myelin (red) and background (black). All SEM and TEM samples shown here are cropped to 512×512 pixels. SEM patches have a pixel size of 0.1 μm, while TEM patches have a pixel size of 0.01 μm (see section "Pipeline overview").

## Pipeline overview

The pipeline of *AxonDeepSeg* is composed of four steps: data preparation, learning, evaluation and prediction. Figure 2 illustrates each step.

In the data preparation step, raw microscopy images and corresponding axon/myelin labels are resampled to a common resolution space: 0.1 μm per pixel for SEM and 0.01 μm for TEM. These values are based on preliminary results and on the typical resolutions provided by each of these imaging systems. Resampled samples are divided into patches of 512×512 pixels due to memory constraints. This size was chosen to have around 15-75 axons per patch. Traditional pre-processing was applied patch-wise, including standardization and histogram equalization (not shown in figure 2 for clarity). For learning, the patches and corresponding labels were randomly split and then considered either for the training or for the validation sets (training/validation split of approximately 70/30%). For evaluation, full test images were randomly selected.

In the learning step, the training/validation dataset is fed into the network. Once the trained model is obtained, performance is evaluated on the test dataset (evaluation step). Finally, the trained model can be used for inference on new microscopy images (prediction step). The images are resampled to the pixel size of the model, divided into patches of 512×512 pixels, segmented, stitched to the native size, and resampled to the native resolution. Note that bilinear interpolation was used during the resampling steps.

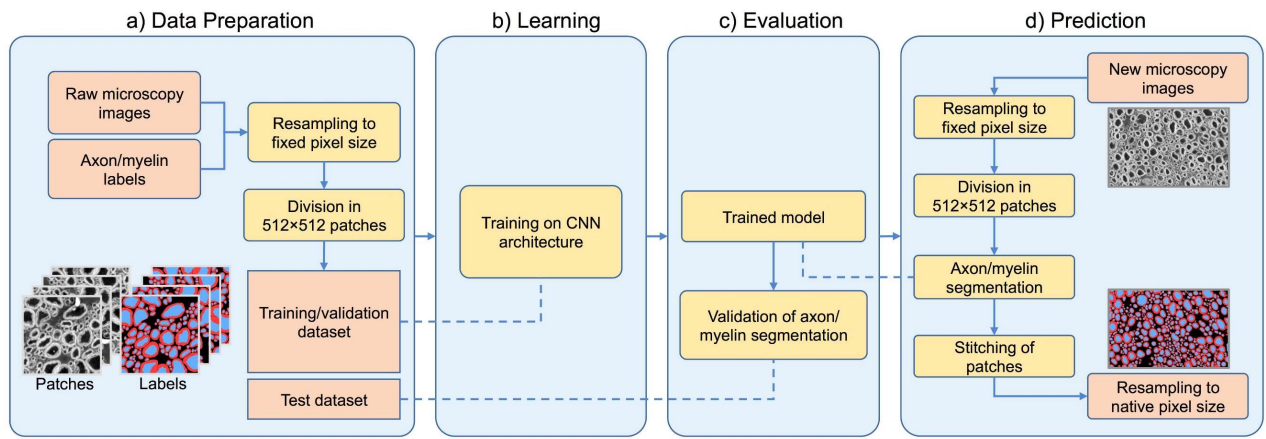

**Figure 2:** Overview of the *AxonDeepSeg* pipeline. During the data preparation step (a), microscopy samples and corresponding gold standard labels are resampled to have a common pixel size (0.1 μm for the SEM model, 0.01 μm for the TEM model), divided into 512×512 patches, and split into training/validation sets. The neural network is trained during the learning step (b) on the training/validation dataset. When the model is trained, performance is assessed on a test dataset (evaluation step (c)). For prediction (d), the new microscopy image to be segmented is first resampled to the working pixel size of the network, divided into 512×512 patches and analysed with the trained model. Segmented output patches are then stitched together and resampled back to the native pixel size.

## Architecture of the network

The architecture is inspired by the original U-Net model[23], combining a contracting path with traditional convolutions and then an expanding path with up-convolutions. Figure 3 illustrates the network architecture. The convolutional layers in the first block use 5×5 kernels, while the convolutional layers on remaining blocks use 3×3 kernels. The SEM network has 3 convolutional layers per block, while the TEM network has 2 convolutional layers per block. These decisions were based on preliminary optimizations (see section "Hyperparameter optimization"). In the contracting path, convolutions of stride 2 are computed after the last convolutional layer of each block to reduce the dimensionality of the features. Each strided convolution layer has a corresponding up-convolution layer in the expansion path in order to recover the localization information lost during the contraction path. Up-convolutions were computed by bilinear interpolation followed by a convolution. The merging of the context and localization information is done by concatenating the features from the contracting path with the corresponding ones in the expansion path. The number of features (channels) is doubled after each block, starting from 16, and then decreased at the same rate during the expansion path. All activation functions in the convolutional layers are rectified linear units (ReLU)[27]. The last layer before the prediction is a softmax activation with 3 classes (axon, myelin and background). The SEM and TEM networks have a total of 1,953,219 and 1,552,387 trainable parameters, respectively.

## Data augmentation strategy

A data augmentation strategy was used on the input patches in order to reduce overfitting and improve generalization[16,19,23]. The strategy includes random shifting, rotation, rescaling, flipping, blurring and elastic deformation[28]. Table 2 summarizes the data augmentation strategy and the corresponding parameters.

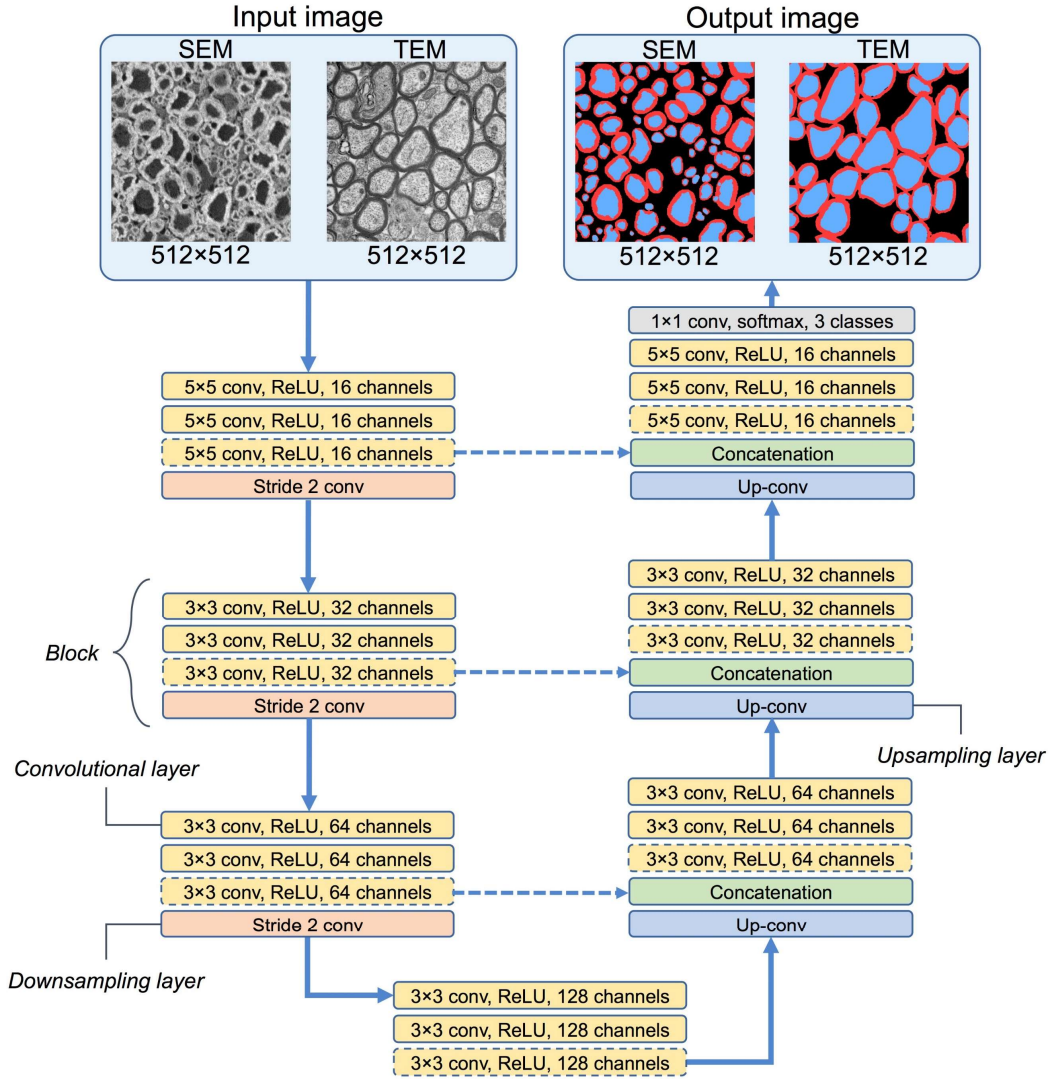

**Figure 3:** Architecture of the convolutional neural networks designed for the segmentation of SEM and TEM images. For the SEM model, 3 convolutional layers are used at each block, while only 2 convolutional layers are used for the TEM model. Convolutional layers in dashed lines are removed for the TEM model. All activation functions used are rectified linear units (ReLU). Strided convolutions are used to downsample the features during the contraction path (left), while up-convolutions are used to recover the localization during the expansion path (right). Features of the contraction path are merged with features of the expansion path to combine localization and context (illustrated by the concatenation step). The pixel-wise classification is done by a 3-class softmax.

# Training procedure

For the training phase, we used a starting learning rate of 0.001 on which we applied a polynomial decay[29] with a power of 0.9. The decay length was 200 epochs, after which the training stopped. We shuffled the samples list at the beginning of each epoch and used a batch size of 8 patches of 512×512 pixels. We have also implemented batch normalization[30] before each activation. The momentum was exponentially decayed from 0.7 to 0.9. This was done to enable a quicker convergence at the beginning of the training by keeping few samples for the batch normalization, while ensuring a stable training at the later epochs. A dropout[31] rate of 0.25 is used in the convolutional layers to reduce the risk of overfitting and improve generalization. The network was trained with the Adam optimizer[32]. We minimized a spatially-weighted multi-class cross-entropy loss. Spatial weights were used to correct class imbalance. The training phase took 86 minutes on an NVIDIA P100 GPU.

| Data augmentation strategy | Description |
| --- | --- |
| Shifting | Random horizontal and vertical shifting between 0 and 10% of the patch size, sampled from a uniform distribution. |
| Rotation | Random rotation, angle between 5 and 89 degrees, sampled from a uniform distribution. |
| Rescaling | Random rescaling of a randomly sampled factor between 1/1.2 and 1.2 |
| Flipping | Random flipping: vertical flipping or horizontal flipping. |
| Blurring | Random blurring: gaussian blur with the standard deviation of the gaussian kernel being uniformly sampled between 0 and 4. |
| Elastic deformation | Random elastic deformation with uniformly sampled deformation coefficient $\alpha=[1-8]$ and fixed standard deviation $\sigma=4$. |

**Table 2:** Data augmentation strategy used in *AxonDeepSeg*. Shifting, rotation, rescaling, flipping, blurring and elastic deformation were applied to training patches in order to reduce overfitting and increase variability.

# Inference procedure

During the inference step, we split the original images into patches of size 512×512 pixels. To overcome border issues (i.e. partial axons at edges not being properly identified as axons), the output segmentation mask is cropped around a smaller patch. Thus, patches overlap by *d* pixels to cover the entire image, as illustrated in Figure 4. Based on preliminary optimizations, the default value *d* was set to 25.

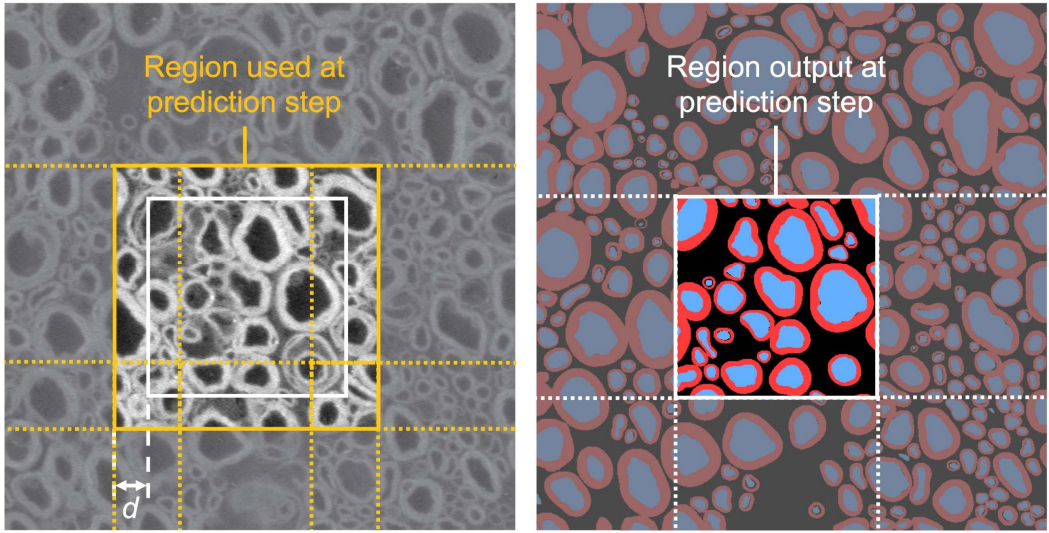

**Figure 4:** Overlapping procedure during inference. To avoid border effects during prediction, inference is run on the orange square, but only the white square is output. The algorithm iterates by shifting the inference window by the size of the white square. The overlap default value *d* was set to 25.

# Hyperparameter optimization

We used different grid searches in order to set the value of the hyperparameters with respect to the accuracy and error on the validation set. The following architecture parameters were optimized at the same time: number of layers, number of filters and convolutional kernel size. The starting learning rate and the batch normalization momentum were also optimized jointly using a grid search, as they both have an effect on the time the model takes to converge and the stability of the validation metrics (based on our experiments). We then jointly optimized the batch normalization momentum and the decay period of the momentum.

# Evaluation method

For testing, the following metrics were computed: the Dice values (axon and myelin) and the pixel-wise accuracy to assess the quality of the segmentation, and the sensitivity and precision to assess the capability to detect true axonal fibers and avoid false axonal fibers.

## Segmentation metrics

To assess the quality of the segmentation we used the Dice coefficient. For two binary images *A* and *B*, the Dice coefficient is defined as:

$$Dice = \frac{2(A \cap B)}{|A| + |B|} \quad (1)$$

where *A∩B* is the intersection between the two images (i.e. number of pixels that are true in both images), $|A|$ is the number of pixels that are true in image *A*, and $|B|$ is the number of pixels that are true in image *B*. The Dice coefficient is computed separately for axon and myelin segmentations, between the prediction and the gold standard masks.

Furthermore, the pixel-wise accuracy is evaluated in order to get a combined assessment of axon-myelin segmentation. The pixel-wise accuracy is computed as the ratio between correctly classified pixels (i.e. axon pixel classified as axon, myelin pixel classified as myelin, background pixel classified as background) and the total number of pixels in the test sample.

## Detection metrics

To assess the performance of myelinated fiber detection, we computed the sensitivity and precision based on axon objects, using the positions of the centroids. Knowing the number of true positives (TP, axons present in both the prediction and the gold standard mask), false positives (FP, axons present in the prediction, but absent in the gold standard mask) and false negatives (FN, axons present in the gold standard mask, but absent in the prediction), we can compute the sensitivity (true positive rate) and the precision (positive predictive value) with the following equations:

$$TPR = \frac{TP}{TP + FN} \quad (2)$$

$$PPV = \frac{TP}{TP + FP} \quad (3)$$

# Data availability

A part of the datasets generated during and/or analysed during the current study are available in the *White Matter Microscopy Database* repository (https://osf.io/yp4qg/). The remaining datasets are available from the corresponding author on reasonable request.

# Results

## Segmentation

Segmentation was evaluated on SEM (rat and human spinal cords) and TEM (mouse splenium and macaque corpus callosum) samples. Segmentation and gold standard masks for both axons and myelin sheaths are displayed on Figure 5. Table 3 lists validation metrics computed on the segmentation outputs: axon Dice, myelin Dice, pixel-wise accuracy, sensitivity and precision.

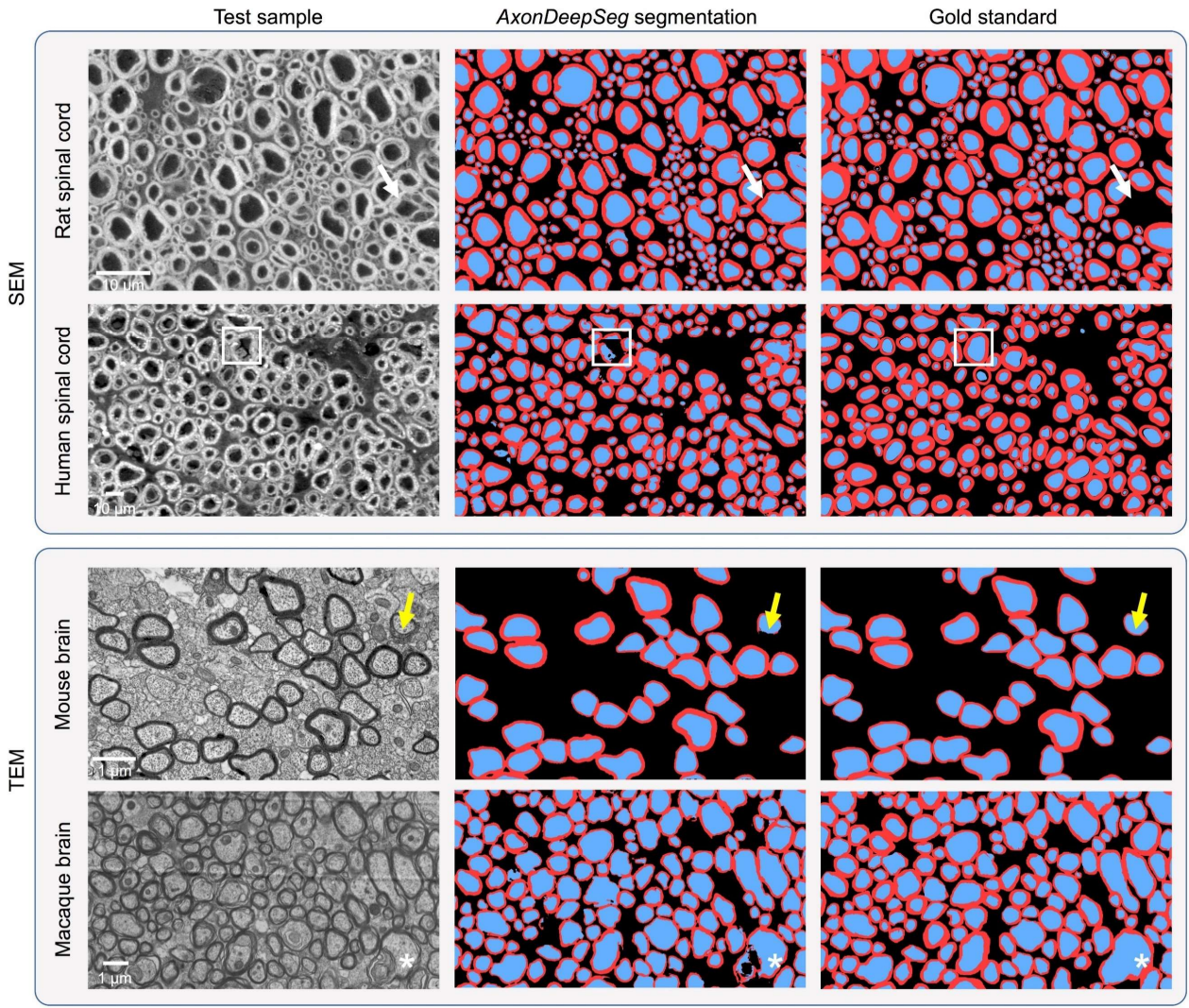

**Figure 5:** Example of segmentation results on SEM and TEM images on a variety of species. The corresponding gold standard segmentation is shown on the right. Overall, the agreement is good. Few discrepancies are noticeable, notably caused by ambiguous/untypical myelin structure (white arrows and white asterisks), inhomogeneous myelin thickness (yellow arrows) and untypical axon intensity (white squares). Some of these discrepancies could potentially be solved using post-processing methods.

To demonstrate the utility of *AxonDeepSeg* for large scale microscopy, segmentation of axon/myelin was performed on a full rat spinal cord SEM (cervical level). Segmentation masks (axons in red, myelin sheaths in blue) are displayed on Figure 6, along with a zoomed window of a small region for better visualization.

| Modality | Model | Test sample(s) | Axon Dice similarity | Myelin Dice similarity | Pixel-wise accuracy | Sensitivity | Precision |
|---|---|---|---|---|---|---|---|
| SEM | Trained on rat samples | Rat 1 | 0.9089 | 0.8193 | 0.8510 | 0.9699 | 0.8468 |
| | | Rat 2 | 0.9244 | 0.8389 | 0.8822 | 0.9876 | 0.7987 |
| | | Human | 0.8089 | 0.7629 | 0.8114 | 0.9300 | 0.7306 |
| TEM | Trained on mice samples | Mice | 0.9493 | 0.8552 | 0.9451 | 0.9597 | 0.9647 |
| | | Macaque | 0.9069 | 0.7519 | 0.8438 | 0.9429 | 0.8129 |

**Table 3:** Summary of performance metrics on test samples, for both SEM and TEM models. The SEM model was trained on rat spinal cord samples, and evaluated on rat and human spinal cord samples, while the TEM model was trained on mice brain samples, and evaluated on mice and macaque brain samples. For each sample, axon Dice, myelin Dice, pixel-wise accuracy, sensitivity and precision were computed. Axon and myelin Dice measure the similarity between the axon/myelin segmentation masks and the gold standard. Pixel-wise accuracy is a measure of the ratio of correctly classified pixels. Sensitivity and precision values are an indication of the capability to detect true axonal fibers and to avoid segmentation of false axonal fibers. Note that for the mice, 24 samples of the same size were used: performance metrics shown are means between all samples.

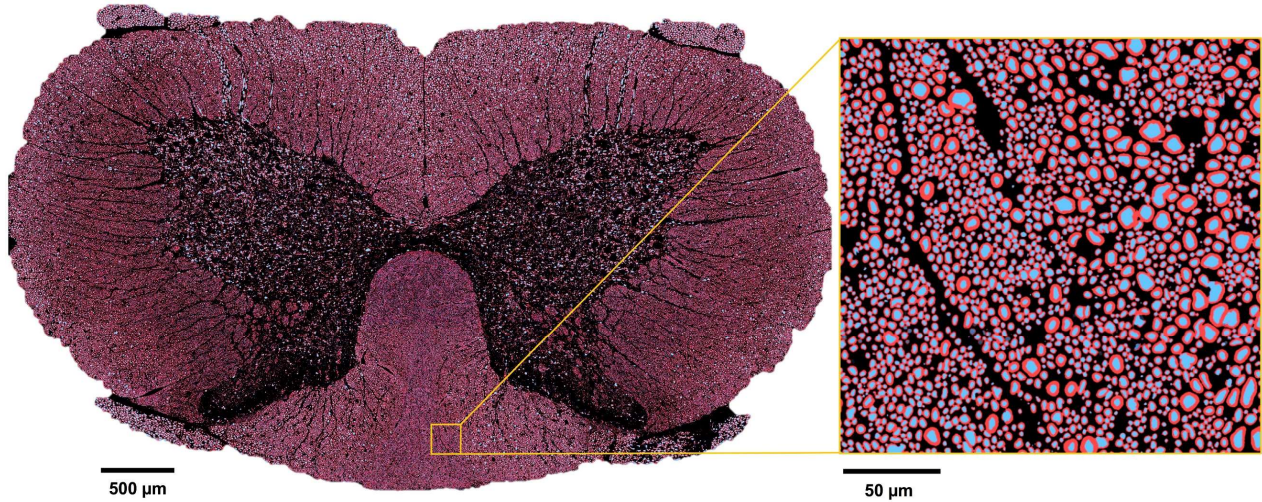

**Figure 6:** Full slice of rat spinal cord showing segmented axons (blue) and myelin sheaths (red). The zoomed panel illustrates the segmentation performance and sensitivity to fiber size: the left half of the panel contains smaller axons (mean diameter around 1.75 µm) while the right half contains larger axons (mean diameter around 2.5 µm).

## Morphometrics extraction

As a proof-of-concept, morphometric statistics were extracted from a full spinal cord of rat using *AxonSeg*[15]. The segmented rat spinal cord shown in Figure 6 was downsampled to 50×50 µm² in order to generate maps of density (e.g., axon and myelin density). The following aggregate metrics were computed:

- Axon diameter mean and standard deviation: arithmetic mean and standard deviation of the distribution of equivalent axon diameters (computed for each axon object as $\sqrt{4*Area/\pi}$);
- Axon density: number of axons per mm²;
- Axon volume fraction (AVF): ratio between area of axons and total area of the region;
- Myelin volume fraction (MVF): ratio between area of myelin and total area of the region;
- G-ratio: ratio between axon diameter and myelinated fiber (axon + myelin) diameter, which can be estimated with the following formula[7]: $\sqrt{1/(1+MVF/AVF)}$.

A binary mask was used to only keep white matter pixels. Results are displayed in Figure 7. Obtained metrics were compared with references of the white matter tracts of the rat spinal cord[33–35].

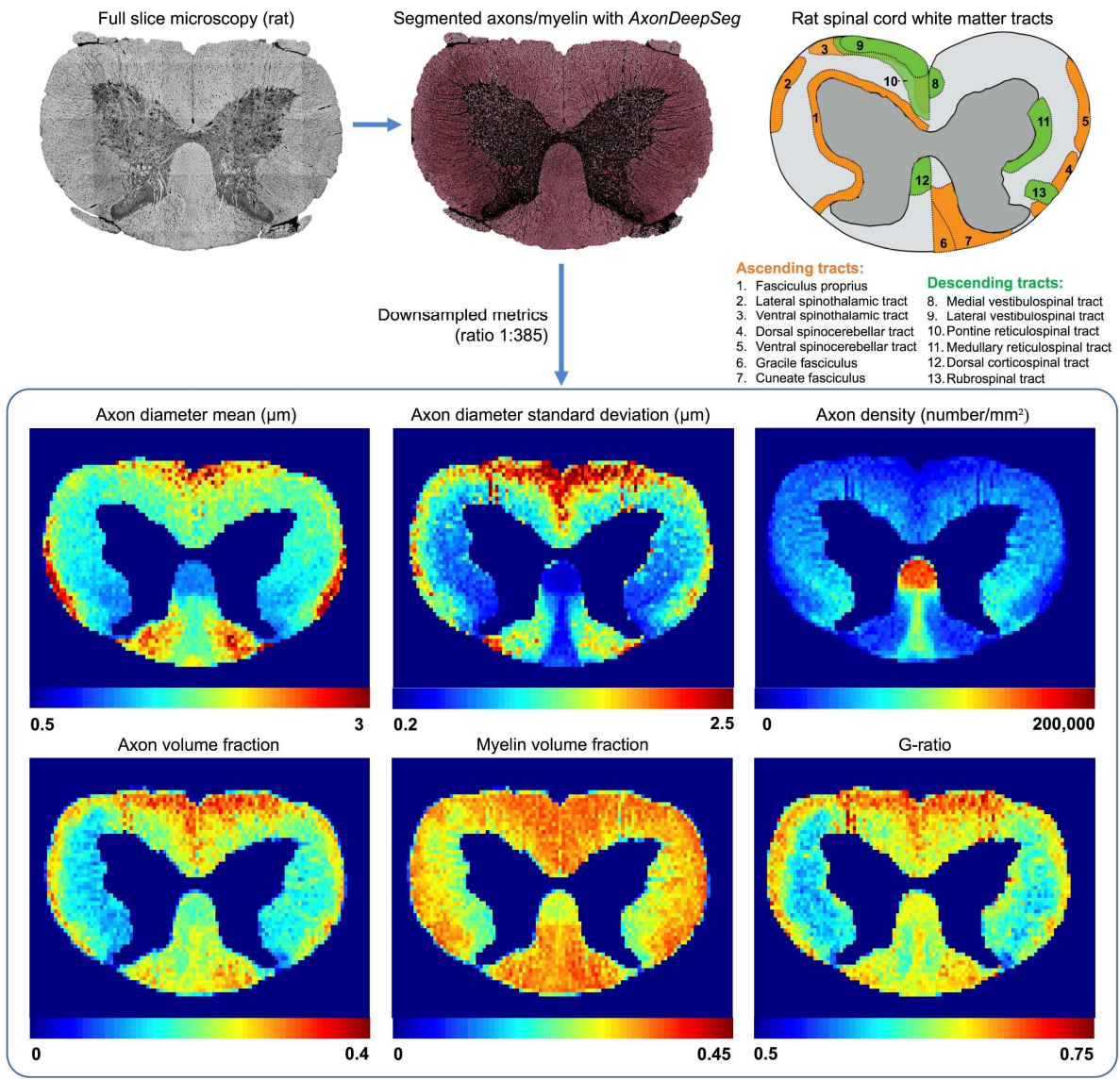

**Figure 7:** Distribution maps of axon diameter mean and standard deviation, axon density, axon volume fraction, myelin volume fraction and g-ratio in a full rat spinal cord slice (cervical level). The SEM slice was segmented with *AxonDeepSeg*. The aggregate metrics of the white matter were generated by downsampling the axon/myelin segmentation masks to a 50×50 µm² resolution. A schematic diagram of the main ascending and descending tracts of the white matter in the rat spinal cord based on the literature[33–35] is provided as reference. The distribution maps are in good agreement with known anatomy. In the corticospinal tract (tract #12 of the reference), we observe smaller axon diameters (around 1 µm), very high axon density (around 200,000 axons per mm²) and g-ratio values around 0.6. Larger axons are found close to the spinal cord periphery. See Discussion for comparison with the literature.

# Discussion

This paper introduced *AxonDeepSeg*, a software framework to segment axon and myelin from microscopy data using deep learning. We will now discuss the resampling methodology, the performance on axon and myelin segmentation, the application to metrics extraction, the software, and future perspectives.

## Resampling

We propose a SEM model trained with a resolution of 0.1 μm per pixel, and a TEM model trained with a resolution of 0.01 μm per pixel. At inference, test image is resampled to meet the target resolution of the model. Other training set compositions were explored, with model trained on both SEM and TEM data in order to achieve better generalization. However, a few limitations arose: (i) SEM and TEM images exhibit very different resolution ranges, requiring large resampling factors to find a common resolution space; (ii) SEM and TEM modalities capture different microstructure/textures of the tissue (for instance, TEM microscopy can capture subcellular microstructure details of the axon); (iii) preliminary results of model simultaneously trained on SEM and TEM led to lower performance when compared to modality-specific models.

## Performance metrics

The SEM model trained on rat microscopy was able to achieve a pixel-wise accuracy between 85% and 88% on the rat test samples, while the pixel-wise accuracy on human test sample was 81% (see table 3). In both test sets, sensitivity was high (>93%), indicating good capability to detect true positive axons. Lower performance metrics obtained in the human test set are expected, as the human sample used exhibits different contrast/quality/noise properties when compared to the rat training set. In the case of the TEM model, performance was very high on the mice test samples: pixel-wise accuracy higher than 94%, both sensitivity and precision around 96%. Good performance was also observed on the macaque test. Note that myelin sheaths of the macaque test sample are slightly underestimated when compared to the gold standard segmentation. In both models and all test samples, computed myelin Dice was lower than axon Dice. This could be explained by the fact that myelin objects have two interfaces: boundary ambiguity between myelin and axon, and boundary ambiguity between myelin and background. Therefore, the myelin Dice is affected by two types of myelin misclassifications: myelin pixel classified as axon or myelin pixel classified as background.

Overall, these results suggest that the trained SEM and TEM models are robust to a variety of species and contrast changes and can generalize well, given that the lowest pixel-wise accuracy observed was 81%. Similar work done on optical microscopy data [25] have achieved a maximal pixel-wise accuracy of 82%. As pointed out in figure 5, most pixel misclassifications are due to ambiguous/untypical axon and/or myelin structure or intensity distribution. Note that these discrepancies could possibly be solved by implementing post-processing methods based on mathematical morphology or conditional random fields.

## Morphometrics extraction

Morphological metrics were extracted from a full rat spinal cord slice at the cervical level (see figure 7). The metrics resulting from the segmentation are overall consistent with the known anatomy. The ventral spinothalamic tract (#3 in tract reference of figure 7) contains the largest axons[33,35], while higher density and smaller axons are observed in the corticospinal tract (#12 in tract reference)[34,35]. Furthermore, the spinocerebellar tracts (#4 and #5 in tract reference) are mostly composed of large diameter fibers[33]. We also observe that axons in the cuneate fasciculus (#7 in tract reference) are larger than those found in the gracile fasciculus (#6 in tract reference), which is also in agreement with the literature[36]. G-ratio ranges between 0.5 and 0.75, which is in agreement with other rat microstructure studies[37]. Overall, concordance of metrics obtained with literature shows that *AxonDeepSeg* can serve as a tool to document distribution and size of myelinated fibers in microscopy samples.

## Software

*AxonDeepSeg* is coded in Python and based on the *TensorFlow* deep learning framework. It can currently run on Linux and Mac OS X systems. Segmentation inference can be done on standard CPU computers at reasonable computational time. For instance, the segmentation of the full rat slice (figure 6) took about 5 hours in a Mac laptop (2.9 GHz). The code is available as open source in GitHub (https://github.com/neuropoly/axondeepseg) and an intuitive documentation is provided (https://neuropoly.github.io/axondeepseg/). A Binder link and a simple Jupyter notebook are available for getting started with *AxonDeepSeg*.

# Future perspectives

The use of ensemble techniques, which consist of combining multiple neural network models, can potentially increase performance metrics. However, its drawback is that it increases computational time at inference. Another possible approach is to use transfer learning[38] in order to obtain better generalization in new imaging modalities even when having a small training set. A partially trained model can be used as starting point for the training of another model of different modality. Note that *AxonDeepSeg* has been trained and tested on healthy tissues. It would be interesting to assess its performance on demyelinated microscopy samples, in which myelin sheaths might present smaller thickness and different morphology.

Even though current models are already performant, our long-term goal is to continuously improve these models by adding more training data from collaborators in order to improve generalization. Another objective is to build segmentation models for other modalities, such as optical microscopy and Coherent Anti-Stokes Raman spectroscopy (CARS). This vision is supported by the recent initiative of creating a White Matter Microscopy Database[39], which provides to the community an open access microscopy data and associated labeled gold standard. We encourage people to share their data for fostering the development of performant segmentation methods.

# Acknowledgements

The authors would like to thank Ariane Saliani and Tanguy Duval for helping with the acquisition of SEM data, Dr. Hugues Leblond for providing the human spinal cord sample, Nafisa Husein and Harris Nami for helping with the gold standard labelling of samples, Drs. Adriana Romero Soriano and Yoshua Bengio (MILA - Montreal Institute for Learning Algorithms), and Dr. Robert Brown (McGill - Montreal Neurological Institute) for fruitful discussions on the network design, Drs. Nikola Stikov and Jennifer Campbell for sharing TEM data of macaque, and Dr. Els Fieremans for sharing TEM data of mice. The authors would also like to thank Compute Canada and Calcul Québec for access to computation units and the "NVIDIA Corporation" for offering a Tesla GPU.

# Author contributions

AZ, MW and JCA wrote the paper. AZ, MW, VH, PLA and CSP designed and developed the software. JCA supervised the project and provided expert guidance. All authors reviewed the manuscript.

# Additional information

**Competing financial interests**: the authors declare no competing financial interests.